\documentclass[letterpaper]{article}

\usepackage{natbib,alifeconf}  
\usepackage{microtype}
\usepackage{graphicx}
\usepackage{subfig}
\usepackage{booktabs} 
\usepackage{enumerate}
\usepackage{multicol}
\usepackage{hyperref}
\usepackage{xcolor}

\usepackage{algorithm}
\usepackage[noend]{algpseudocode}

\algdef{SE}[CLASS]{Class}{EndClass}%
   [2]{\textbf{class} \textproc{#1}\ifthenelse{\equal{#2}{}}{}{(#2)}}%
   {\algorithmicend\ \algorithmicfunction}%
    
\algtext*{EndClass}

\title{Neural MMO v1.3: A Massively Multiagent Game Environment \\ for Training and Evaluating Neural Networks}
\author{Joseph Suarez$^{*1}$, Yilun Du$^{*1}$, Igor Mordach$^{\#1}$, Phillip Isola$^{*1}$\\
\mbox{}\\
$^*$Massachusetts Institute of Technology, Cambridge, MA 02139 \\
$^\#$Google Brain, Mountain View, CA, 94043 \\} 

\begin{document}
\maketitle

\begin{abstract}
Progress in multiagent intelligence research is fundamentally limited by the number and quality of environments available for study. In recent years, simulated games have become a dominant research platform within reinforcement learning, in part due to their accessibility and interpretability. Previous works have targeted and demonstrated success on arcade, first person shooter (FPS), real-time strategy (RTS), and massive online battle arena (MOBA) games. Our work considers massively multiplayer online role-playing games (MMORPGs or MMOs), which capture several complexities of real-world learning that are not well modeled by any other game genre. We present Neural MMO, a massively multiagent game environment inspired by MMOs and discuss our progress on two more general challenges in multiagent systems engineering for AI research: distributed infrastructure and game IO. We further demonstrate that standard policy gradient methods and simple baseline models can learn interesting emergent exploration and specialization behaviors in this setting.\footnote{This publication summarizes v1.0-v1.3. * and \# denote current author affiliations. Work for v1.0 \citep{DBLP:journals/corr/abs-1903-00784} was performed at OpenAI and work for v1.3 was performed at MIT}
\end{abstract}
\vspace{-5mm}

\begin{figure*}[ht!]
\begin{center}
\resizebox{\linewidth}{!}{
\includegraphics{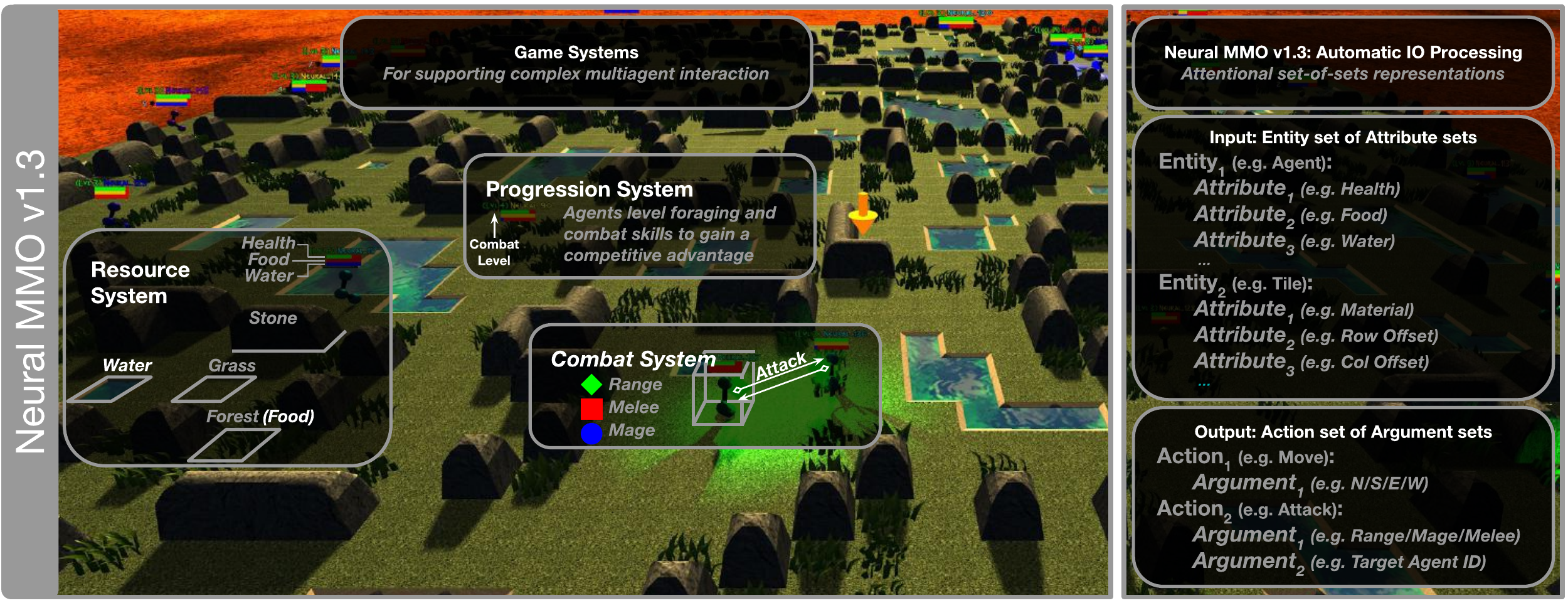}
}
\end{center}
\vspace{-4mm}
\caption{Neural MMO is a massively multiagent environment for AI research. Agents compete for resources through foraging and combat. Observation and action representation in local game state enable efficient training and inference. A 3D Unity client provides high quality visualizations for interpreting learned behaviors. The environment, client, training code, and policies are fully open source, officially documented, and actively supported through a live community Discord server.}\label{fig:cent}
\label{fig:header}
\vspace{-2mm}
\end{figure*}

\section{Introduction} 
From arcade to FPS to RTS and MOBA, the use of increasingly complex game environments has accelerated progress in deep reinforcement learning (RL) in recent years  \citep{silver2016mastering, baker2019emergent, OpenAI_dota, alphastarblog, jaderberg2018human}. MMOs are possibly the most complex class of games in the entertainment industry and are a natural a next step in this progression. They simulate self-contained macrocosms with large, variable numbers of players, user-driven economies, team-oriented strategizing, and realistic long-term planning over hundreds to thousands of hours of persistent gameplay. We argue that, among all game genres in the entertainment industry, MMOs produce a style of in-game learning that comes closest to learning in the real world.

Our eventual goal is capable artificial intelligence within a full MMO through continuous environment development and iterative research upon it. Neural MMO seeks to capture the most important properties of the base game genre in a more simplified setting. In the process of working towards this objective, we encountered two major research engineering problems surrounding infrastructure and IO. This work shares our solutions to these problems in the context of Neural MMO as general methods that enable scalable multiagent learning in complex environments. The key contributions of this publication are:

\begin{enumerate}
\item Neural MMO as a fully open-source and actively supported environment for multiagent research
\vspace{-2mm}
\item Pretrained policies with the associated distributed training code and utility libraries for reproducibility [\href{https://youtu.be/DkHopV1RSxw}{\color{blue}{Video}}]\footnote{We have linked a one minute video of trained policies. Full address in case of hyperlink errors: https://youtu.be/DkHopV1RSxw}
\vspace{-2mm}
\item Stand-alone scalable infrastructure and performance logging for massively multiagent environments
\vspace{-2mm}
\item Stand-alone methods for interfacing with complex observation and action spaces in multiagent environments 
\end{enumerate}

\textbf{Infrastructure:} Modern deep RL frameworks place assumptions on the environment that are untrue or computationally inefficient in large multiagent systems. This situation has forced us to reexamine standard RL infrastructure according to computation placement and communication patterns. We formalize our findings in Ascend, a lightweight wrapper on top of Ray \citep{DBLP:journals/corr/abs-1712-05889}. Ascend provides general abstractions and design patterns for multiagent systems that allow us to implement the full Neural MMO training backend in only a few lines of code.

\textbf{IO Spaces:} Small scale RL environments typically provide input observations as raw data tensors and output actions as low-dimensional vectors. More complex environments may contain variable length observation and action spaces with mixed data types: these IO spaces cannot interface with standard architectures that expect fixed length tensors.\footnote{Rendering does not solve the issue. See The IO Problem, below} Our solution to this problem parameterizes the observation space as a set of entities (in turn parameterized by a set of attributes) and automatically generates attentional networks to select variable length action arguments by keying against learned entity embeddings.

\section{Related Work}
While deep reinforcement learning (RL) has recently expanded to include a variety of control focused tasks, games have always been an important research platform~\citep{DBLP:journals/corr/MnihKSGAWR13}. The Arcade Learning Environment (ALE)~\citep{bellemare2013arcade} and Gym Retro~\citep{nichol2018gotta} provide 1000+ limited scope arcade games most often used to test individual research ideas or generality across many games. More recent work has demonstrated success on multiplayer games including the board game Go~\citep{silver2016mastering}, the card game Heads-Up No-Limit Poker ~\citep{DBLP:journals/corr/MoravcikSBLMBDW17}, the Multiplayer Online Battle Arena (MOBA) game DOTA2~\citep{OpenAI_dota}, the Real Time Strategy game Starcraft 2 ~\citep{alphastarblog}, first person Hide and Seek ~\citep{baker2019emergent}, and Quake 3 Capture the Flag ~\citep{jaderberg2018human}. These tasks are difficult and important but are limited to 2-10 players, are episodic with game rounds less than an hour, lack persistence, and lack game mechanics supporting large populations. Our work seeks to expand game AI to MMOs, which do possess these properties. 

``Artificial life" (ALife) aims to model evolution and natural selection in biological life; \citep{langton1997artificial, ficici1998challenges}. Such projects often consider open-ended skill learning ~\citep{yaeger1994computational} and general morphology evolution ~\citep{sims1994evolving2} as primary objectives. Some environments in this space simulate tens to upwards of a million agents ~\citep{lowe2017multi, mordatch2017emergence, bansal2017emergent, lanctot2017unified, yang2018mean, zheng2017magent, jaderberg2018human}. Most such works further focus on learning a specific dynamic such as predator-prey ~\citep{yang2018study} or use hard-coded rewards because they are more concerned with studying emergent behavior than learning it from scratch ~\citep{zheng2017magent}. Our work is based in games rather than biological life. While the real world is substantially more complex, MMOs support similar persistent social dynamics in large agent populations. Unlike the real world, they are efficient to simulate and straightforward to develop; numerous tools and best practices for MMO creators already exist courtesy of the entertainment industry.

\begin{figure*}[htp]
\centering
\includegraphics[scale=0.74]{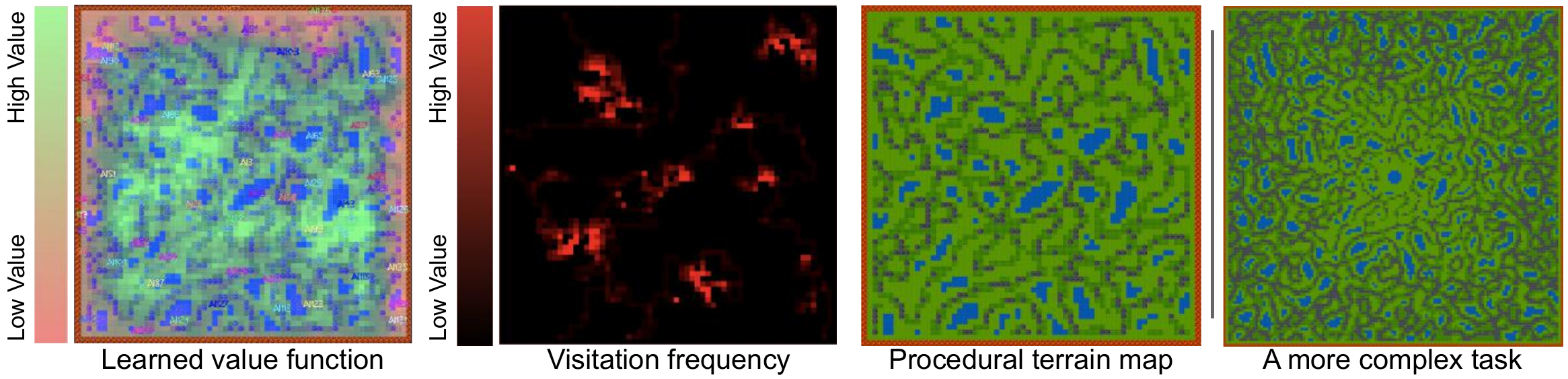}
\caption{Visualizations render the learned value function and agent exploration patterns directly within the game client. In our experiments, we use the map on the left. Procedural generation enables a variety of task complexities: the rightmost map is 2x larger and also manipulates resource distributions to scale difficulty as agents move farther from the center.}
\label{fig:overlay}
\end{figure*}

\section{Neural MMO}
Neural MMO is a massively multiagent environment for artificial intelligence research. Agents forage for resources and engage in strategic combat within a \textit{persistent} game world that is never reset during training. Our environment implements a progression system inspired by traditional MMOs and a full 3D renderer for visualizations (see Figure \ref{fig:header}).

\textbf{Environment Representation:}  Neural MMO is laid out on 2D tile map that is procedurally generated by thresholding a Perlin ridge fractal. For all maps, agents are added (\textit{spawn}) within a designated region at a rate of one per timestep (server \textit{tick}) up to a cap of 128. Agents may move about the grass and forest tiles of the game map, but stone and water are impassible. The map is also surrounded by a lethal lake of lava. Figure \ref{fig:header} shows terrain types, and Figures \ref{fig:header} and \ref{fig:overlay} (left) show the latest version of the game map for which we have successfully trained policies.\footnote{On this map, agents forage inwards from the borders.} The map generation code is configurable to enable a variety of different studies within our environment. For example, Figure \ref{fig:overlay} (right) shows a larger map in which generation has been tweaked to produce easily navigable but resource poor terrain (see Resource System) near the center and mazelike but resource rich farther out.\footnote{On this map, agents forage outwards from the center.}

\textbf{Resource System:} Agents spawn with 10 food, water, and health. At every tick, agents lose one food and one water. All of these values are configurable. If agents run out of food or water, they begin losing health. If agents are well fed and well hydrated (i.e. above half food and water), they begin regaining health. In order to survive, agents must quickly forage for food, which is in limited supply, and water, which is infinitely renewable but only available at a smaller number of pools.\footnote{Agents collect food by moving into a "forest" tile and water by moving adjacent to a "water" tile. Forest tiles turn to grass once consumed, but they regenerate over time. Figure \ref{fig:header} shows tile types} Thus, the objective is simultaneously navigate and forage for food/water in the presence of upward of a hundred agents attempting to do the same.

\begin{figure*}[!htp]
\centering
\renewcommand{\floatpagefraction}{.99}%
\renewcommand{\dblfloatpagefraction}{0.99}
\resizebox{.90\linewidth}{!}{
\includegraphics[]{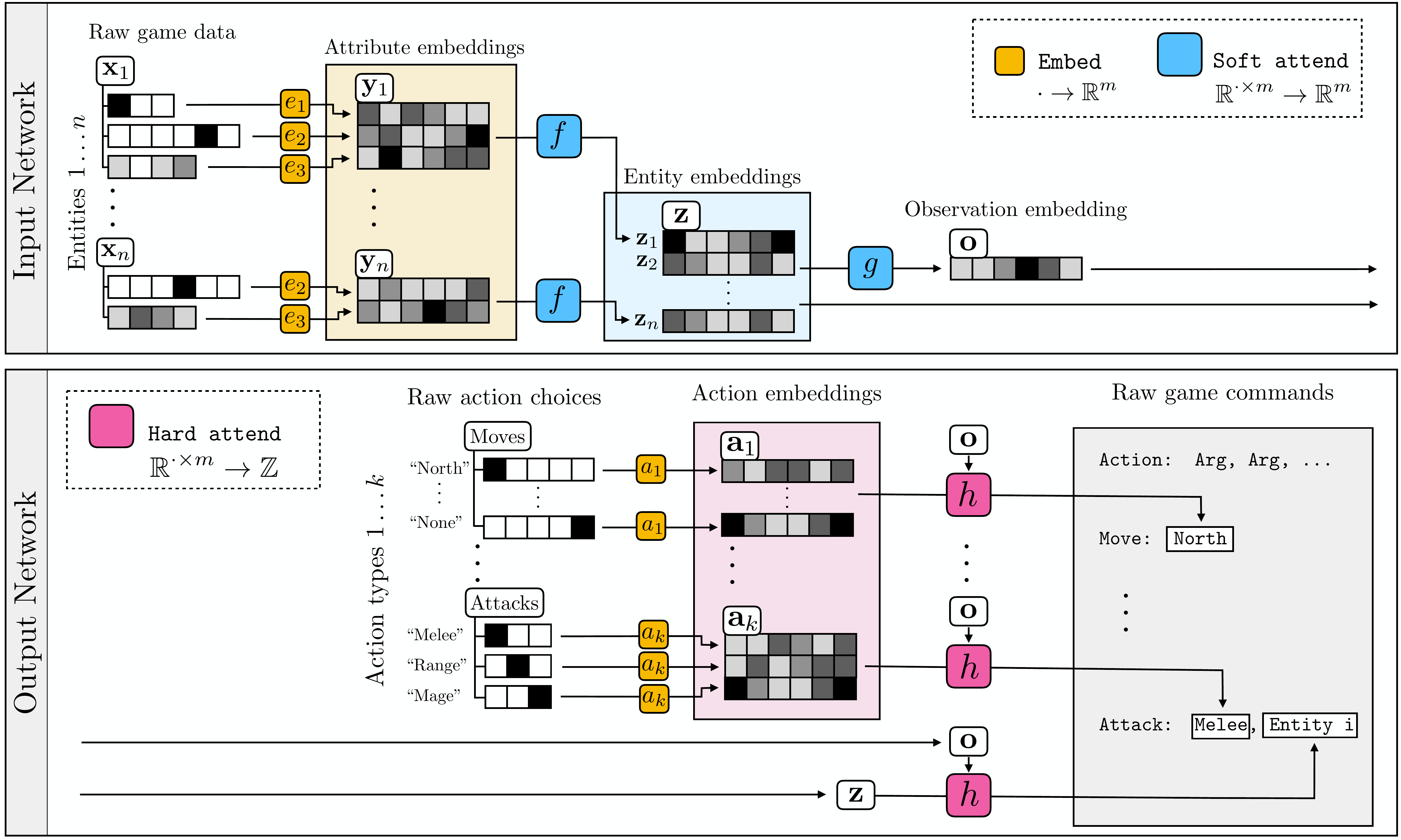}
}
\vspace{-2mm}
\caption{The environment provides local game state as a two-layer observation hierarchy over all nearby entities (tiles and other agents) and their associated attributes. The \textit{input} module applies attention over attributes to produce entity embeddings. A second attentional mechanism is applied over entities to produce a flat observation embedding. The \textit{outputs} module keys the intermediate entity embeddings against the network hidden state to select action-argument pairs with hard attention.}
\label{io}
\end{figure*}

\textbf{Combat System: } Agents can attack each other with three different styles. For flavor, we refer to these as Range, Mage, and Melee. Accuracy and damage are determined by the attack style and the combat stats of the attacker and defender (see Progression System). This system, which also allows agents to pilfer resources from their target, was designed to enable a variety of strategies. For example, Range and Melee attacks are more damaging, but a successful magic attack freezes the target in place temporarily. Agents more skilled in combat can assert map control, locking down resource rich regions for themselves. Agents more skilled in maneuvering can succeed through foraging, using mage attacks to prevent aggressors from closing distance. In short, agents must balance the reduced risk of attempting to forage passively, protecting themselves only when needed, against the greater reward of attacking their neighbors to pilfer their resources and cull the competition. \footnote{Large scale battles at spawn are undesirable in MMOs. For the map used in our baseline (Figure \ref{fig:header}), agents may not attack agents that have recently spawned. For the new map in Figure \ref{fig:overlay}, agents may be attacked only by other agents within a fixed distance of their level (see Progression System). Agents are perfectly safe within a small region of the spawn containing little to no food. As they travel farther, increasingly powerful agents are able to attack them, but more resources are present. This situation adds another risk-reward trade off layer to the environment.}

\textbf{Progression System:} Progress in real MMOs varies on two axes: soft advantage gained through strategic/mechanical talent and hard numerical advantage gained through skill levels/equipment. In Neural MMO, agents progress their abilities through usage. Foraging for food and water grants experience in the respective Hunting and Fishing skills. A higher Hunting level enables agents to gain more from resource tiles and also carry more food. The same is true of Fishing and water tiles, and a similar system is in place for combat. Agents gain levels in Constitution, Range, Mage, Melee, and Defense through fighting other agents. Higher offensive levels increase accuracy and damage. Higher Constitution directly increases an agent's maximum health. Higher Defense decreases the accuracy of opponents' attacks. An overall level is calculated for each agent based off of their combat stats. The global scale of experience awarded for each action is configurable to enable any game progression time scale from a few minutes to thousands of hours. 

\textbf{Front End Client:} The Neural MMO client is written in C\# using Unity3D and follows a design shared by multiple MMOs: while the game state is a 2D tile grid, the client renders in full 3D with smooth animations. As the client is only used for test-time visualization, this approach enables us to achieve far greater computational efficiency than physically simulated environments.\footnote{In our experiments, only 0.5-2.0 percent of total CPU is used to simulate the environment} At the same time, access to a full game engine allows us to maintain game play complexity without sacrificing interpretability or visual fidelity. A unique programmable name is displayed above each agent along with its health, food, and water values. We also display additional status effects, such as whether an agent is frozen and the level range within which it is attackable. Users can click on specific agents to follow them with the camera or view more detailed stats. Our previous THREE.js client contained a number of useful research overlays for visualizing agent exploration behaviors and value maps (Figure \ref{fig:overlay}), which will be ported to the Unity3D client soon.

\section{The IO Problem}
\begin{figure}[t]
\centering
\includegraphics[scale=.50]{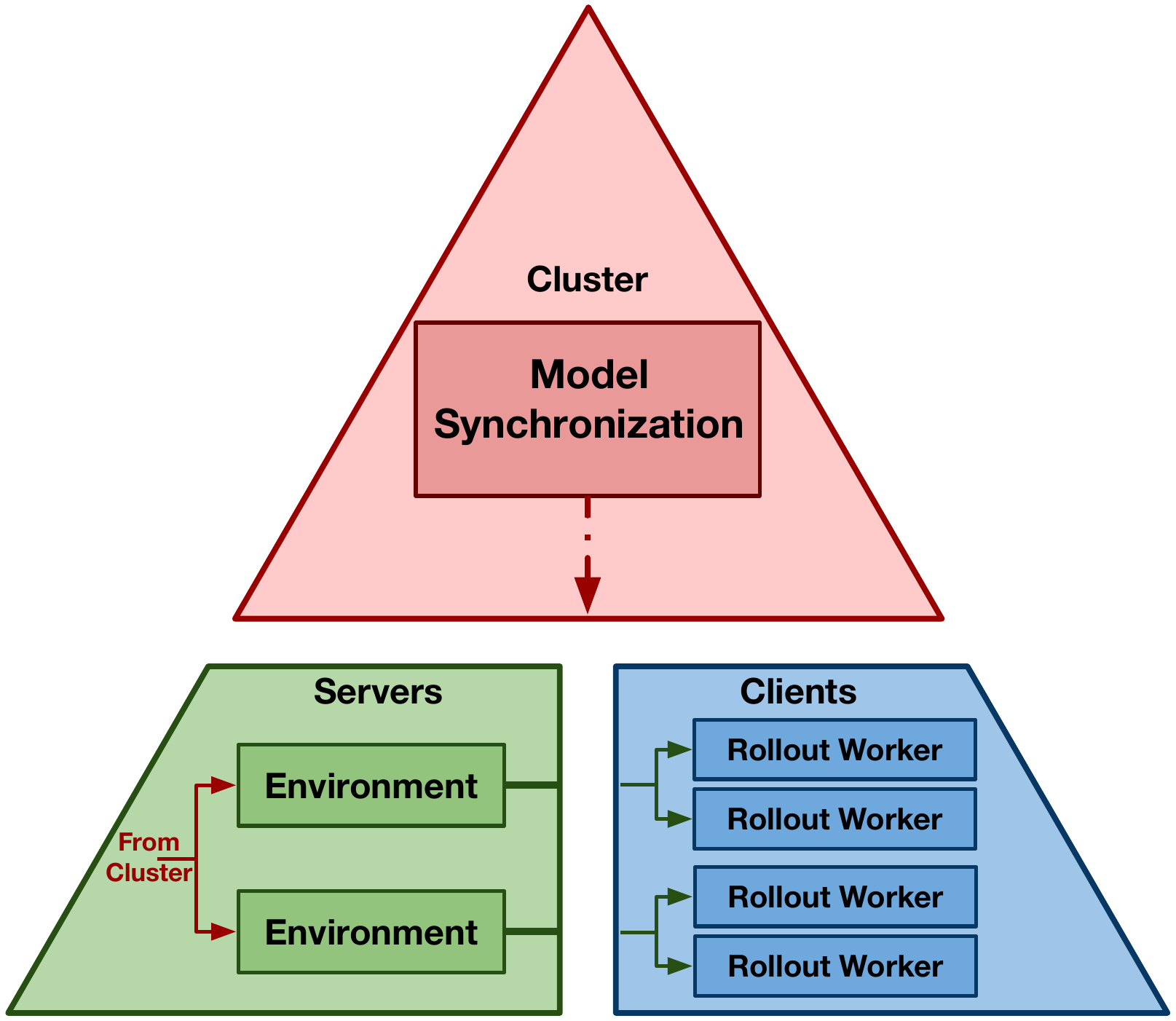}
\caption{Hardware diagram of Neural MMO infrastructure (Algorithm \ref{ascend}) in three Ascend layers. The main difference from Rapid is our placement of the environment on the server rather than the client.}
\label{infra}
\end{figure}

Most simulated environments used in reinforcement learning offer an interface in which observations are represented as raw tensors and actions are selected by sampling from flat logits: it is a simple task to input observations into the network and convert network outputs into actions. However, in more complex environments, observation and actions spaces may be complex, of variable size, and hierarchical. As demonstrated by recent scale up work on DoTA, the exact structure of these IO spaces can require complex environment specific architectures simply to enable natural representations of observations and actions. We refer to these issues collectively as \textit{the IO problem}. This work presents pluggable modules that makes interfacing with Neural MMO almost as simple as working with a traditional game environment. Instead of hand engineering a network to match the particular IO spaces of Neural MMO, we have taken the opportunity to design a more generic framework for input and output processing applicable to a broad class of environments. Our solution implements a substantially general automatic network generation process (see Figure \ref{io}) that can be implemented as a layer on top of existing environments. While our work is in an interpreted language, we will refer to \textit{compile time} and \textit{run time} to denote static and instance specific operations.

\textbf{Notes:} One tempting solution is to simply render the environment and give agents the same interface as humans. However, rendering complex environments is almost always much more computationally expensive than simulating agent policies. Without rendering, agents no longer can use a mouse and keyboard -- there is nothing to click on. Our work bypasses this human interface and allows agents to interact with the raw game API.

\subsection{The Input Problem}
\textbf{Terminology:} We define local game state by the set of observable objects or \textit{entities} (e.g. agents and tiles), each of which is parameterized by a number of local properties or \textit{attributes} (e.g. health, food, water, etc.).

\begin{algorithm}[t]
\caption{The Ascend API provides three abstractions over standard distributed infrastructure.}\label{ascend}
\begin{algorithmic}[1]
\Class{Ascend}{}
\Function{distribute}{arguments, shard=None}
	\State asyncHandles = List()
	\State shardedArguments = Shard(arguments, shard)
	\For {worker in remote workers}
		\State handle = worker.step(shardedArguments)
		\State asyncHandles.append(handle)
	\EndFor
	\State \textbf{return} asyncHandles
\EndFunction 
\Function{synchronize}{asyncHandles}
	\State remoteReturns = List()
	\For {handle asyncHandles}
		\State data.append(await(handle))
	\EndFor
	\State \textbf{return} data
\EndFunction 
\Function{step}{arguments, shard=None}
	\State asyncHandles = distribute(arguments, shard)
	\State \textbf{return} synchronize(asyncHandles)
\EndFunction
\EndClass
\end{algorithmic}
\end{algorithm}

\textbf{Assumptions:} We assume that the attributes for each entity type are declared at compile time. This implies that it is possible to build a static tree of attribute data types without querying specific entities at run time. Our work currently supports continuous and discrete attributes, though it is possible to support additional data types in the future. We assume reasonable estimates of each attribute's scale. For discrete values: a lower bound on the minimum value and an upper bound on the maximum value. For continuous values: a rough estimate of the mean and standard deviation. Note: estimates aid in input normalization but are not strictly necessary. It is also possible to collect them empirically at runtime.

\textbf{Solution:} Figure \ref{io} describes our solution to the input problem. At compile time, the user specifies embedding functions for each attribute type (e.g. different functions for embedding continuous and discrete values). Our framework uses these function handles and the static attribute tree in order to generate embedding layers $e_1, e_2, ...$ for each attribute. The user also specifies two attention functions, $f$ and $g$ to be used later. At run time, we use the static graph structure and associated layers $e_i$ to embed $x_1, x_2, ..., x_n$, producing fixed length attribute embedding vectors $y_1, y_2, ..., y_n$. To produce a fixed length vector representation $z_i$ for each entity, we aggregate across attributes using the specified attention function $f$. The action network will need these entity embeddings later. To obtain a fixed length representation $o$ of the entire observation, we aggregate across observed entities $z_1, z_2, ..., z_n$ using the second specified attention function $g$. Finally, we return the flat observation embedding and the variable sized lookup tensor of entity embeddings.  

\subsection{The Output Problem}
\textbf{Terminology:} We define agent decision space by a set of action-argument pairs. Actions are callable function references that the environment can invoke on the associated argument list in order to execute an agent's decision. For example: Move $\to$ [North] or Attack $\to$ [Melee, Agent ID]

\textbf{Assumptions:} We assume that the set of actions and their argument types are declared at compile time. Our work currently supports discrete, and entity valued arguments, though it is possible to support additional data types in the future.

\textbf{Solution:} Figure \ref{io} describes our solution to the output problem. At compile time, the user specifies a hard attentional architecture $h$ and output networks (unembedding layers) for each argument type. Using the static action set, our framework then generates decision layers for each argument. At run time, we convert the hidden state of the main network into an action + argument-list pair. To do so, we first embed all arguments to produce fixed length vector representations $a_1, a_2, ..., a_k$. For entity values arguments, we simply copy the $z_1, z_2, ...$ entity representations from the input network. We then compare these embeddings to the hidden state using the attentional function $h$ to produces hard attentional choices over arguments. Finally, we match the selected embedding to the corresponding argument game object.

\section{Distributed Infrastructure}
Modern reinforcement learning infrastructure makes a number of assumptions about the underlying task. Most notably, it has become standard practice to centralize all policy computation and optimization on a centralized GPU server, using a bank of remote CPU clients to produce training data by simulating many environment instances in parallel. This practice is most often implemented as a simple broadcast and aggregate operation in raw MPI. We will refer to this common usage as the MPI approach, though MPI itself is a much more general framework. OpenAI has recently proposed an alternative infrastructure configuration, Rapid, which they used to train agents to play DoTA ~\citep{OpenAI_dota} above human skill. This was one of the largest (and possibly the largest) reinforcement learning project undertaken at the time, and communication latency became a problem. Rapid remedies this issue by collocating agent policies with dedicated GPU rollout workers that simulate full agent trajectories independently. Separate dedicated GPU optimizer servers aggregate experience in large batches from many rollout workers and environments, thus removing the need for constant high-latency communication among remote hardware.

\begin{figure}[t]
\centering
\includegraphics[scale=0.355]{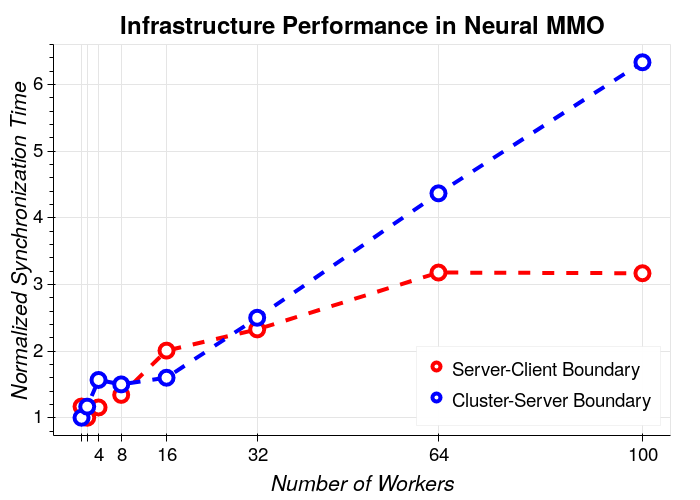}
\caption{Neural MMO sync. times scale linearly with \# cluster workers and sublinearly with \# server workers.} \label{tab:infra_table}
\label{tab:scale}
\end{figure}

We argue that neither MPI nor Rapid scale well in massively multiagent settings. In their work on DoTA, OpenAI used approximately 100 environments per GPU optimizer server. Classic MMOs such as Ultima Online and Runescape supported thousands of players per live server on 90s to early 2000s hardware. In a deep learning setting, scaling to this number of agents per environment decreases the number of environments that each server can support. MPI and Rapid both assume a one-to-many ratio of optimizers to environments; real MMOs by necessity operate using a many-to-one ratio. Instead of supporting hundreds of environments per server, suddenly one (or multiple) servers are required to support each environment. The bandwidth optimization of Rapid over MPI is an orthogonal improvement that does not alter this one-to-many ratio assumption. Our work implements MMO style infrastructure in deep learning to perform distributed training in Neural MMO.

\subsection{Ascend Distributed API}

Ascend is a protocol that models infrastructure and associated logging at an arbitrary hardware layer. We implement Ascend as a lightweight wrapper around Ray, a popular general purpose distributed computing library. Our API provides a single eponymous Ascend object which specifies synchronous and asynchronous communication interfaces to the previous and next hardware layers. Stacking three such layers produces the cluster-server-client architecture needed for Neural MMO in only a few lines of specialized code.

Algorithm \ref{ascend} details the API. Ascend provides three core subroutines: distribute, synchronize, and step. The distribute function invokes all workers in the next layer asynchronously. It returns awaitable handles and supports argument sharding across clients. The synchronize function waits for all remote clients, as invoked by distribute, to terminate and aggregates their returns. The step function provides a synchronous remote interface by simply calling distribute followed by synchronize. The user specifies behavior at each hardware level by overriding these three methods.

\begin{algorithm}[tb]
   \caption{Neural MMO training logic for one game tick. Note that we abstract rollout collection.}
   \label{alg:train}
\begin{algorithmic}
   \For{\textbf{each} environment server}
      \If{number of agents alive  $<$ spawn cap}
         \State spawn an agent
      \EndIf
      \For{\textbf{each} agent}
         \State i $\leftarrow$ population index of the agent
         \State Make observation $o_t$, decide action $\pi_i(o_t) \to a_t$
         \State Environment processes $a_t$ and computes $r_t$
         \If{agent is dead}
            \State remove agent
         \EndIf
      \EndFor
      \State Update environment state $s_{t+1} \rightarrow f(s_t, a_t)$
   \EndFor
   \State Perform a policy gradient update on policies $\pi \sim {\pi_1, \ldots, \pi_N}$ using $o_t$, $a_t$, $r_t$ from all agents across all environment servers
\end{algorithmic}
\end{algorithm}

\subsection{Neural MMO}
Ascend is a general framework that can be used to implement the computational models of MPI and Rapid in only a few lines. Figure \ref{infra} details our specific usage of Ascend in Neural MMO, which has three layers. The cluster layer controls a single node used for top level experiment management, logging, and model update synchronization across a bank of servers. Each server simulates a Neural MMO environment instance, performs various observation preprocessing and action postprocessing, and communicates with a bank of clients. Each client manages a dynamic, variable length subset of the corresponding server's agents. This includes batching observations, running the policy, collecting rollouts, and performing backpropagation. We focus on reinforcement learning in this work, but our framework also supports other approaches, such as genetic algorithms.

\begin{figure*}[t]
\centering
\includegraphics[scale=0.7]{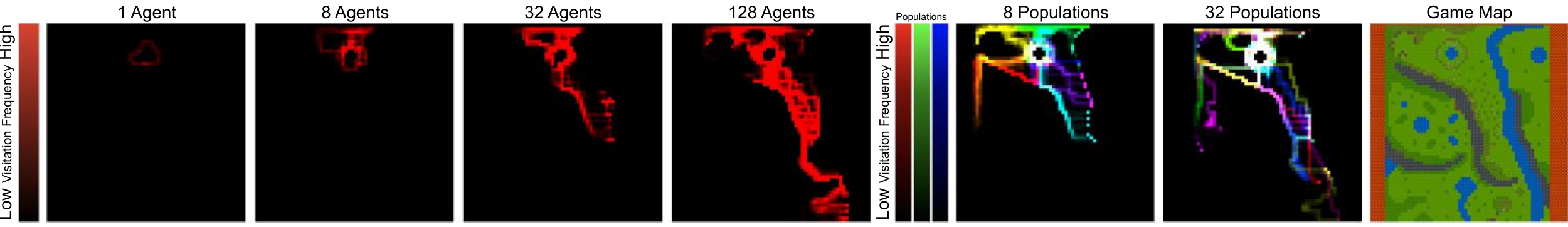}
\caption{Visitation frequency overlaid over the game map. (Left) Population size magnifies exploration: agents spread out to avoid competition. (Right) Populations count (unshared policies, separated by color) magnifies niche formation (8 populations) until all exploration strategies are saturated (32 populations).}
\label{experiments}
\end{figure*}

\subsection{Scale}
We evaluate the performance of Ascend in Figure \ref{tab:scale} by considering synchronization times at two boundaries: cluster-server and server-client. We vary the number of cluster workers (environment servers) and server workers (policy clients) independently. For all trials, we evaluate synchronization time for one gradient step over a batch of size 4096 agent decisions per environment. This experiment was conducted on a small local cluster of 10 machines with 10 usable cores each.

The cluster receives a fixed length vector of gradient updates from each server. Thus, linear synchronization time is the ideal result for the cluster-server boundary, with bandwidth as the limiting factor. At the server-client boundary, the same amount of data is shared across all clients regardless of their number. As such, the ideal result is constant synchronization time independent of the number of clients. The empirical results shown in Figure \ref{tab:scale} are consistent with the expected trends. For both synchronization boundaries considered, results are noisy only at very small scale (one machine). As we scale from 10 to 100 CPU cores, the cluster-server synchronization time converges to linear performance. Server-client synchronization appears to plateau with a roughly constant overhead factor of 3. Some deviation from ideal is to be expected. We expect this factor could be reduced through better load balancing and serialization: clients do not receive perfectly equal splits of the observations, and serializing many small data packets to be sharded across several clients is less efficient than serializing a few larger packets.

\section{Experiments}

The main contributions of this work are our environment and the associated general purpose multiagent systems. However, we also create an end-to-end training pipeline on Neural MMO in order to test our infrastructure and IO. In the Policy and Training sections below, we train populations of agents to perform basic foraging and combat behaviors on the latest version of our environment mechanics and the map generation shown in Figures \ref{fig:header} and \ref{fig:overlay} (left). The Emergent Behavior section describes interesting qualitative properties of multiagent interaction discovered using Neural MMO v1.0.

\subsection{Policy}

We define our architecture by the functions $f, g$, and $h$ in Figure \ref{fig:overlay}. For the function $f$, which attends over attribute embeddings, we use one layer of single-headed scaled dot product attention ~\citep{DBLP:journals/corr/VaswaniSPUJGKP17}. The funtion $g$ applies attention over agent embeddings and convolution over tile embeddings to produce a flat observation embedding. In the output network, agents submit one movement and one attack action per game tick. The function $h$ that is used to select arguments first applies a small fully connected network to the hidden state and argument embeddings. We then use dot product similarity followed by a softmax to compute a distribution over action arguments. Sampling from this distribution produces the final argument selections.

\subsection{Reward Formulation and Training}

Agents are controlled by policies parameterized by neural networks. Each agent makes a partial observation $o_t$ of the game state $s_t$ and follows a policy $\pi(o_t) \to a_t$ in order to make action(s) $a_t$. We maximize a return function $R$ over trajectory $\tau = (o_t, a_t, r_t, ..., o_T, a_T, r_T)$. Neural MMO provides a custom reward shaping API, but for simplicity we use a discounted sum of survival rewards of form $[0, 0, ..., -1]$: $R(\tau) = \sum_t^T \gamma^tr_t$ where $\gamma = 0.95$, $T$ is the time at death, and reward is $r_{t=T}=-1; r_{t \not = T} = 0$. Each rollout corresponds to an agent lifetime. As described in Algorithm \ref{alg:train}, we sample agent policies from eight different populations and train with policy gradients ~\citep{sutton:nips12} using a learned value baseline ~\citep{NIPS1999_1786}. The policy weights $\pi$ are shared only within each population, excepting embedding weights, which are shared across all populations.

The v1.3 baseline uses a batch size of 16k actions \footnote{our v1.0 experiments used a noisier policy gradients formulation and required a batch size of 265k actions}. The cluster aggregates gradients across all servers and broadcasts policy updates back to the clients, ensuring weights are never stale. We clip gradients to a maximum absolute value of 5.0 and update the network using Adam with learning rate 3e-4 and weight decay 1e-5. We observe that agents learn basic foraging and combat behaviors and release trained policies.
\vspace{-1mm}
\subsection{Emergent behavior}
These experiments were performed on an earlier version of the environment, but the core game remains unchanged, and our open source release includes a branch with this version of the environment for reproducibility. Key differences from the current environment include map generation, lack of a progression system, and slight mechanical tweaks.

\subsubsection{$N_{ent}$: \#Agents Magnifies Exploration}

The left half of Figure \ref{experiments} compares map coverage vs. population size. In the natural world, competition between animals can incentivize them to spread out in order to avoid conflict. We observe that overall exploration (map coverage) increases as the number of concurrent agents increases. Agents learn to explore only because the presence of other competing agents provides a natural incentive for doing so.

\subsubsection{$N_{pop}$: \#Populations Magnifies Niche Formation}

The right half of Figure \ref{experiments} compares map coverage vs. number of populations (agents with unshared weights). We find that, given a sufficiently large and resource-rich environment, different populations of agents tend to separate to avoid competing with other populations. The real world often rewards masters of a single craft more jacks of all trades. Figure \ref{experiments} suggests that specialization to particular regions of the map increases as number of populations increases. We believe this indicates that the presence of other populations force agents to discover a single advantageous skill or trick: increasing the number of populations results in diversification to separable regions of the map. As entities cannot out-compete other agents of their own population (because they share the same policy), they tend to seek large areas of the map that contain enough resources to sustain their entire population.

\section{Discussion and Limitations}
We have made significant progress towards enabling seamless reinforcement learning research on massively multiagent environments, but much is left to be done. In this section, we describe the portions of our work that we believe could benefit from additional generalization.

\subsection{IO}
The object centric model of observation space enables a robust attentional mechanism over attributes and entities to create a natural and general mechanism for flattening a complex local game state. It further enables direct object queries in the action space by keying against entity embeddings. However, we found it difficult to train attention layers over a large number of tile embeddings. This is the reason we used a convolutional layer over tile embeddings in our experiments. While we expect attention to be trainable with some additional network tuning, memory remains an issue. Attention is quadratic in the number of entities, and agents observe a 15 by 15 crop of tiles, or 225 total. One potential solution is to factorize the attentional mechanism. Standard attention prepossesses the input $x$ with three linear layers, $Q$, $K$, and $V$. Instead, one could, for example, replace each of $K, Q$, and $V$ with $k$ linear layers. By taking max pooled average over $k$ projections of $x$ along the dimension corresponding to the number of entities, it is possible to reduce the memory required for activations to $O(k^2 + n)$.

Our output module is a more temporary solution: it works for the level of complexity in modern environments, but it will require support for additional data types to be applicable in something as complex as a full MMO. However, we emphasize that reinforcement learning environments, including Neural MMO, are not yet complex enough for this to be a concern. That is, this discussion will not be relevant until after a few iterations of environment development.

\subsection{Infrastructure}
The current environment version is in fact amenable to Rapid style infrastructure. However, neither MPI nor Rapid style infrastructure will scale well once we begin increasing agent population sizes. This factored into our decision to invest in infrastructure early. Another factor in our decision was a quirk of reinforcement learning regarding memory. People play MMOs for hundreds to thousands of hours, and decisions made fifty hours into the game impact play five hundred hours later. Because of reward discounting, we need to keep full trajectories for all living agents in memory. We tried multiple schemes for reducing memory overhead, including recomputing the forward pass, but we found this to be too complex and computationally expensive to merit further consideration. Possible alternative approaches include the trajectory segment processing formulation used in OpenAI Five, off-policy methods that consider single transitions, and evolutionary methods that do not keep trajectories in memory at all. Intrinsic reward methods could also be useful to reduce the effective required time scale.

\section{Conclusion}

Neural MMO has been in development for over two years and is now fully open source with dedicated setup, documentation, and tutorial pages, an active Discord community support server with over 80 current members, and major updates every 3-4 months. We plan to support Neural MMO as a robust platform for multiagent research at small and large scale and will continue its development going forward. The environment has also served as a useful case study in multiagent systems, which has allowed us to address infrastructure and IO problems in this emerging new research space. We hope that our solutions will prove useful to others developing massively multiagent systems.

\section{Acknowledgments}
This project had been made possible by the combined effort of many contributors over the last two years. Joseph Suarez is the primary architect and project lead. Igor Mordatch managed and advised the project at OpenAI. Phillip Isola advised the project at OpenAI and resumed this role once the project residence shifted to MIT. Yilun Du assisted with experiments and analysis on v1.0. Clare Zhu wrote about a third of the legacy THREE.js web client that enabled the v1.0 release. Finally, several open source contributors have provided useful feedback and discussion on the community Discord as well as direct bug fixes and features. Additional details are available on the project website. \\

\noindent This work was supported in part by:\\
\indent Andrew (1956) and Erna Viterbi Fellowship\\
\indent Alfred P. Sloan Scholarship (G-2018-10127)

\footnotesize
\bibliographystyle{apalike}
\bibliography{nmmo_v1-3}

\end{document}